%% file: main.tex
\documentclass[sigconf]{acmart}
\usepackage{amsmath}
\usepackage{ulem}

\AtBeginDocument{%
  }

\setcopyright{acmlicensed}
\copyrightyear{2018}
\acmYear{2025}
\acmDOI{XXXXXXX.XXXXXXX}
\acmConference[KDD'25 FedKDD workshop]{Federated Learning for Data Mining and Graph Analytics }{August 4,
  2025}{Toronto, ON, Canada}
\acmISBN{978-1-4503-XXXX-X/2018/06}

\usepackage[dvipsnames,table]{xcolor}
\usepackage{algorithm}
\usepackage{algpseudocode} 

\usepackage{enumitem}

\usepackage{subcaption}

\usepackage{adjustbox}
\usepackage{caption}
\newcommand{\J}[1]{\textcolor{black}{#1}}

\newcommand{\Eason}[1]{\textcolor{black}{#1}}
\begin{document}

%\title{Hierarchical Federated Unlearning for Large Language Models}
\title{Hierarchical Federated Unlearning for Large Language Models}
\author{Yisheng Zhong}
\email{yzhong7@gmu.edu}
\affiliation{%
  \institution{George Mason University}
  \city{Fairfax}
  \state{VA}
  \country{USA}
}

\author{Zhengbang Yang}
\email{zyang30@gmu.edu}
\affiliation{%
  \institution{George Mason University}
  \city{Fairfax}
  \state{VA}
  \country{USA}
}

\author{Zhuangdi Zhu}
\email{zzhu24@gmu.edu}
\affiliation{%
  \institution{George Mason University}
  \city{Fairfax}
  \state{VA}
  \country{USA}
}

\renewcommand{\shortauthors}{Yisheng Zhong et al.}


\vspace{0.1in}
\begin{abstract}

%\J{ The storyline that we discussed and agreed on:}

%\J{LLM unlearning is important, yet faces two challenges: practical unlearning requests may be 1) heterogeneous and continual, instead of coming in one go, and 2) unlearning tasks may be data sensitive and decentralized.  This leads to inter-domain and intra-domain unlearning interference, which further amplifies the dilemma of unbalanced forgetting and retaining performance of LLMs.}
%\J{To address these two challenges, we propose a federated learning approach that is efficient and data privacy preserving.  Specifically, to decouple the unlearning from retraining, we xxx (split and unlearn).To address the heterogeneous unlearning tasks, we xxx (hierarchical aggregate). Comprehensive experiments on xx, xx, and xx showed that xxxx.}

Large Language Models (LLMs) are increasingly integrated into real-world applications, raising concerns about privacy, security and the need to remove undesirable knowledge. Machine Unlearning has emerged as a promising solution, yet faces two key challenges: (1) practical unlearning needs are often continuous and heterogeneous, and (2) they involve decentralized, sensitive data with asymmetric access. These factors result in inter-domain and intra-domain interference, which further amplifies the dilemma of unbalanced forgetting and retaining performance.
In response, we propose a federated unlearning approach for LLMs that is scalable and privacy preserving. Our method decouples unlearning and retention via task-specific adapter learning and employs a hierarchical merging strategy to  mitigate conflicting objectives and enables robust, adaptable unlearning updates.
%we design a hierarchical merging strategy to effectively accommodate heterogeneous and decentralized unlearning tasks. Additionally, we decouple unlearning from retraining through task decomposition for more robust model performance. %to independently manage forgetting and knowledge retention.
%
Comprehensive experiments on benchmarks of WMDP, MUSE, and TOFU showed that our approach effectively handles heterogeneous unlearning requests while maintaining strong LLM utility compared with baseline methods.\footnote{Code will be published at: \url{https://anonymous.4open.science/r/Unlearning-B493}}

\end{abstract}

\begin{CCSXML}
<ccs2012>
<concept>
<concept_id>10010147.10010257.10010258</concept_id>
<concept_desc>Computing methodologies~Learning paradigms</concept_desc>
<concept_significance>500</concept_significance>
</concept>
</ccs2012>
\end{CCSXML}

\ccsdesc[500]{Computing methodologies~Learning paradigms}

\keywords{LLM unlearning, Federated unlearning, Hierarchical model merging, Task heterogeneity, Unlearning-utility trade-off}


\maketitle

\section{Introduction}

Large Language Models (LLMs) have shown remarkable capabilities in generating content that resonates with human knowledge. 
While LLMs are increasingly integrated into real-world applications, growing concerns about privacy and copyright protection~\cite{voigt2017gdpr, pardau2018ccpa} have called for mechanisms to actively remove undesirable knowledge previously learned by these models.
Accordingly, \textit{Machine Unlearning} has emerged as a promising technique to tackle this need without retraining the model from scratch~\cite{yao2024large,nguyen2024surveymachineunlearning, xu2023machineunlearningsurvey}. 

%%% Challenge of unlearning itself
{Prior LLM unlearning methods assume access to both the data to be forgotten and a representative subset of the original training data to retain general knowledge~\cite{liu2025rethinking}. 
They commonly optimize on a dual objective with competing components: one optimizing on the unlearning data, and the other the retaining data~\cite{maini2024tofutaskfictitiousunlearning},  yet still face trade-offs including  \textit{incomplete unlearning} of the target content and \textit{over-forgetting} of retention knowledge.
%
%%% Challenge of multi unlearning 
Moreover, in realistic scenarios}, unlearning requests can arise {\it externally}, {\it continuously}, from {\it decentralized sources}, ranging from white-box model users and intellectual property owners to red-teamers, who often have individual unlearning needs but are reluctant to reveal the sensitive unlearning data.%, especially given that the data designated for forgetting is inherently private. 

In response, we investigate a practical paradigm of  {\textit{Federated Unlearning for LLMs}}, which is built upon a federated learning mechanisms~\cite{mcmahan2017communication} to tackle  LLM unlearning with asymmetric data access: the party initiating the unlearning request ({\it client}) may lack access to {\it retaining data}, while the LLM provider ({\it server}) is restricted from directly accessing the unlearning data due to privacy or security regulations.
This scheme enables decentralized unlearning requests  without requiring centralized data collection.

%%% Challenge of naiive federated unlearning 
Despite its innovative framework,  our empirical study shows that a naive federated mechanism alone often results in performance degradation, either degrading unlearning performance or at the sacrifice of retention utility drop, which reflects the persistent challenge in existing unlearning and becomes more pronounced given decentralized and heterogeneous unlearning requests.

\begin{figure}[h]
% \vspace{-0.1in} 
  %\centering
  \includegraphics[width=0.5\textwidth]{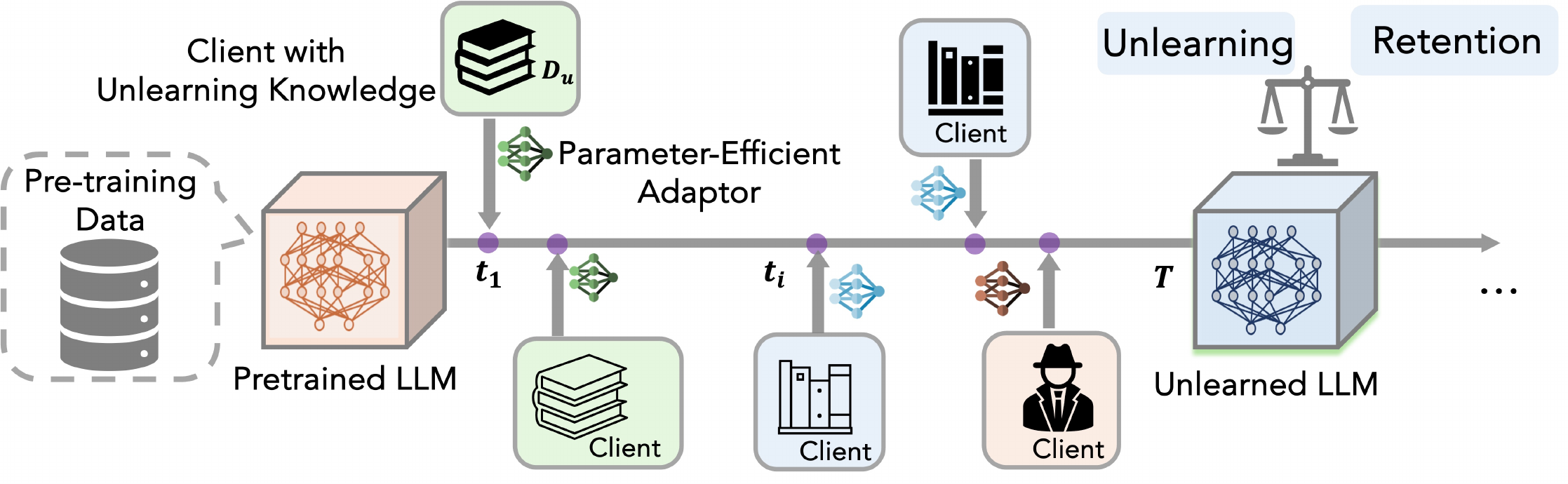}
  % \vspace{-0.1in}
  \caption{\begin{small} Our method tackles heterogeneous, decentralized LLM unlearning requests without transmitting the sensitive unlearning or retention data.\end{small}}
  \label{fig:overview}
\vspace{-0.25in}
\end{figure}

%%% Method
%%% 
To address these entangled challenges, we propose a systematic algorithm and framework, named Federated UnLearning Merge (FULM), to synergize the decentralized unlearning requests continuously directed to an LLM server, as shown in Figure \ref{fig:overview}.
FULM decouples the unlearning and retention objectives and instead integrates retention signals during federated aggregation to preserve useful knowledge in the LLM.  By analyzing client updates, FULM identifies disparities within task adapters regarding parameter distributions, magnitudes, and directions, and applies hierarchical aggregation strategies tailored to these patterns.

%We evaluate our proposed hierarchical federated unlearning approach by integrating three unlearning benchmarks for LLMs: WMDP, TOFU, and MUSE. Results demonstrate that, compared to traditional FL or unlearning baselines, FULM is effective in removing undesirable knowledge while preserving LLM utility and is scalable to accommodate dynamic unlearning tasks within the complex LLM structures, all without the need to aggregate either unlearning or retention data, thus offering an accountable and privacy-preserving approach to LLM knowledge governance.
We evaluate our approach on three LLM unlearning benchmarks: WMDP, TOFU, and MUSE. Results show that given dynamic unlearning tasks, FULM outperforms existing baselines of unlearning or FL merging, and can effectively remove target knowledge while preserving model utility, thus offering an accountable and privacy-preserving approach to LLM knowledge governance.
% \section{Method}

% Our federated unlearning framework utilizes LoRA-based task vectors provided by individual clients to protect local data privacy and enhance data transfer efficiency. We observed that task vectors from iid data demonstrate high distribution consistency, while non-iid task vectors exhibit substantial variability. Consequently, we designed two distinct merging strategies:

% \textbf{iid data}: A voting-based merging approach is employed, aggregating similar task vectors to ensure robust generalized model performance.

% \textbf{non-iid data}: Direct arithmetic addition ensures comprehensive incorporation of diverse task vectors, preventing potential omissions.

% The merging operation can be formally represented as:
% \begin{equation}
% \theta' = \theta + \text{Vote}(\Delta \theta_i^a + \dots + \Delta \theta_{i+n}^a) + \Delta \theta_b + \Delta \theta_c + \dots + \Delta \theta_k
% \end{equation}

%\vspace{-0.1in}
\section{Related Work}

\textbf{Machine unlearning} has recently been applied to LLMs to address security and privacy concerns ~\cite{yao2024largelanguagemodelunlearning, 7163042, 234843, mantelero2013right}. %Please cite a few more papers such as surveys  
Early approaches primarily involve \textit{gradient ascent} that fine-tunes the model to increase prediction error on the forget set~\cite{jang2022knowledgeunlearningmitigatingprivacy}.
%, which often results in instability and unintended degradation of unrelated knowledge~\cite{ilharco2023editingmodelstaskarithmetic}. 
%
Different efforts have been proposed to improve unlearning stability. 
%these issues.
%
Negative Preference Optimization (NPO)~\cite{zhang2024negativepreferenceoptimizationcatastrophic} reframes unlearning as a \textit{preference alignment} task by treating the forget set as negative examples, \J{requiring curation of paired response data regarding forgetting knowledge.} %
Task vector methods~\cite{ilharco2023editingmodelstaskarithmetic} subtract the influence of forgetting knowledge from the model by \J{negating the parameters of task adapters related to forgetting knowledge}, while interpolation-based methods like WHP~\cite{eldan2023whosharrypotterapproximate} softly blend target and reinforced models.

To preserve model utility during unlearning, recent work proposed regularization based strategies to maintain performance on a retention set. Gradient Descent on the Retain Set (GDR)~\cite{liu2022continuallearningprivateunlearning, maini2024tofutaskfictitiousunlearning, zhang2024negativepreferenceoptimizationcatastrophic} adds a standard cross-entropy loss over the retain data, while KL Divergence Minimization (KLR)~\cite{maini2024tofutaskfictitiousunlearning, zhang2024negativepreferenceoptimizationcatastrophic} aligns the output distributions of the unlearned and original models. 
These methods improved stability but assume \J{simultaneous access to both the unlearn and the retain dataset}.
\J{A few works explored retaining data-free LLM unlearning at the cost of non-negligible utility degradation~\cite{wang2025llm}}.
%and are not easily extensible to federated or continuous settings. 
%
% Our work addresses this gap by enabling decentralized, task-level unlearning with utility-aware vector aggregation.

\noindent \textbf{Federated Learning and Unlearning:} Federated Learning (FL) is a decentralized machine learning technique that enables clients to collaboratively train models without sharing raw data~\cite{Li_2020,mcmahan2017communication}.
FL was initially proposed for uniform client distributions and evolved to tackle data or systematic heterogeneity~\cite{mcmahan2017communication, 10468591, li2022federated}. 
%
%Nevertheless, its design does not natively accommodate unlearning requests~
\cite{9521274, Liu_2022}. 
A few pioneering works explored federated unlearning on traditional, smaller-scale neural networks for class-wise prediction tasks.
%\cite{9521274, Liu_2022}.
%
Methods include client-side retraining~\cite{9521274}, knowledge distillation~\cite{wu2022federatedunlearningknowledgedistillation}, or model parameter pruning~\cite{wang2022federatedunlearningclassdiscriminativepruning},
which presume access to class labels~\cite{wang2022federatedunlearningclassdiscriminativepruning}, extensive retraining rounds~\cite{Liu_2022}.
%, and may not suffice for LLM unlearning scenarios.

\noindent \textbf{Adapter Merging} for LLMs aims to integrate multiple independently-trained adapters into a single model for multi-task purposes without further fine-tuning.
While Low-Rank Adaptation (LoRA)~\cite{hu2021loralowrankadaptationlarge} has become dominant for parameter-efficient fine-tuning, merging these adapters presents significant challenges due to weight entanglement and interference between task-specific updates~\cite{tang2024parameterefficientmultitaskmodel}.
%has become a dominant parameter-efficient fine-tuning technique for LLMs. However, merging LoRA modules for multi-task performance presents significant challenges due to stronger weight entanglement~\cite{tang2024parameterefficientmultitaskmodel}. 
%Recent studies show that naive LoRA fusion underperforms and call for specialized merging strategies~\cite{tang2024parameterefficientmultitaskmodel, yadav2023tiesmergingresolvinginterferencemerging}. 
%
\textit{\textbf{Task Arithmetic}} (TA)~\cite{ilharco2023editingmodelstaskarithmetic} is a straightforward paradigm for LLM adapter composition by linearly combining fine-tuned LLM adapters, called task vectors. This approach enables merging without retraining, but has been shown to suffer from interference between task-specific updates~\cite{yadav2023tiesmergingresolvinginterferencemerging, daheim2024modelmerginguncertaintybasedgradient}. 
To address this, recent methods include pruning low-magnitude weights~\cite{yadav2023tiesmergingresolvinginterferencemerging}, rescaling sparse vectors~\cite{he2025localizeandstitchefficientmodelmerging}, or aligning models in tangent space to reduce conflict~\cite{ortizjimenez2023taskarithmetictangentspace}.
Specifically, \textbf{TIES}~\cite{yadav2023tiesmergingresolvinginterferencemerging} mitigates this through a three-step process: (1) trimming negligible parameter changes to reduce noise, (2) selecting dominant signs across adapters for each parameter, and (3) averaging updates matching the selected signs. 
Our work connects to prior work by efficiently merging multiple unlearning and retention adapters for balanced model utility, without assuming access to relevant data or post-hoc finetuning.
Our empirical study (Section~\ref{sec:exp}) demonstrated that neither task arithmetic nor the TISE method alone perfectly addresses the heterogeneous unlearning task merging and thus motivated a more tailored merging approach.

%These strategies improve robustness for general model fusion, yet primarily assume full model access and task homogeneity.}

%Crucially, existing TA and LoRA merging techniques focus on task transfer or multi-task generalization and have not been extended to support unlearning scenarios, where model updates may be adversarial, heterogeneous, or privacy-driven. Furthermore, their assumptions of centralized data and synchronous fine-tuning limit their applicability in federated or continual settings, where knowledge must be selectively removed or preserved. 

%To address these gaps, we propose a federated unlearning framework that operates directly on LoRA task vectors, supports heterogeneous forgetting requests, and preserves retention knowledge through distribution-aware vector merging.

\section{Preliminaries}

\textbf{Machine Unlearning.} Machine unlearning is a mechanism to selectively remove undesirable knowledge from a trained model without the need to retrain the model from scratch. Given a model $\theta$, a forget set $\mathcal{D}_u$ containing undesirable knowledge, and a retention set $\mathcal{D}_r$ representing useful knowledge, a representative unlearning objective optimizes the following:
\begin{equation} \label{eq:unlearn-dual}
\min_{\theta'} \mathcal{L}_{\text{unlearn}}(\mathcal{D}_u;\theta') + \lambda \mathcal{L}_{\text{retain}}(\mathcal{D}_r;\theta'),
\end{equation}
where $\lambda$ balances the trade-off between forgetting and retention. Conventional unlearning methods usually implement gradient ascent on $\mathcal{L}_{\text{unlearn}}$, and optionally apply regularization techniques such as KL-divergence minimization to constrain the model parameter divergence from $\theta$ to $\theta'$~\cite{maini2024tofutaskfictitiousunlearning}. Optimizing the above goal requires simultaneous access to both datasets of $\mathcal{D}_u$ and $\mathcal{D}_r$. 

\noindent \textbf{Federated Learning (FL).} 
%FL allows decentralized training of a global model $\theta$ across $K$ clients without sharing client data. 
In a standard FL framework, a client $k \in [K]$ holds a local dataset $\mathcal{D}_k$, which computes a local gradient update $\Delta \theta_k$ based on the global model $\theta$:
%\begin{equation}
$\Delta \theta_k^{(t)} = - \eta \nabla \mathcal{L}(\mathcal{D}_k;\theta^{(t)}),$ 
%_{\theta_k^{(t)}}
%\end{equation}
where $\nabla \theta_k^{(t)}$ is the local model update after training with the global model copy $\theta^{(t)}$ on local data $\mathcal{D}_k$ at communication round $t$, and $\eta$ is the learning rate. The server aggregates gradient updates from different clients to update the global model:
%\begin{equation}
$\theta^{(t+1)} \leftarrow \theta^{(t)} + \frac{1}{K} \sum\limits_{k=1}^{K}\Delta \theta_k^{(t)}.$ 
%\end{equation}
FL typically involves multiple communication rounds $t$ to iteratively improve the global model.
%multiple communication rounds $t$ to allow  to benefit from distributed data while preserving data privacy.
In contrast, we employ a one-shot FL setting and omit frequent parameter exchanges.
%FL inherently preserves data privacy, as only local parameter updates, not raw data, are communicated, though it does not directly accommodate asymmetric unlearning scenarios.

\noindent \textbf{Low Rank Adaptation (LoRA).} Full model fine-tuning on LLMs is computationally expensive and memory-intensive. To address this, LoRA provides parameter-efficient fine-tuning of pretrained LLMs by introducing low-rank matrices. Instead of directly modifying the pretrained weight matrix $W \in \mathbb{R}^{d \times d}$, LoRA defeines the adapted weight matrix $W'$ as:
%\begin{equation}
$W' = W + BA,$ 
%\end{equation}
where $W$ denotes the original frozen pretrained weights, and $B \in \mathbb{R}^{d \times r}, A \in \mathbb{R}^{r \times d}$ are low-rank matrices with rank $r \ll d$. 
LoRA is particularly well-suited for machine unlearning, as the knowledge designated for removal usually occupies a small subspace of the model's overall knowledge and thus can be effectively captured by low-rank adaptations.
%Since both unlearning and retention updates tend to produce low-magnitude parameter changes, LoRA can compactly encode them, making it well-suited for efficient representation and aggregation in our framework.

\iffalse 
\noindent \textbf{adapter Merging} \J{(To be merged with related work)} aims to integrate multiple adapters that were trained independently into a single model for multi-task purposes without further model fine-tuning. Notable approaches include \textbf{TIES}~\cite{yadav2023ties} and \textbf{Task Arithmetic}~\cite{}.
%
The merging process of TIES has three main steps: 1) Trim, where adapter parameters with negligible changes are removed to reduce noise. 2) Elect Sign, where the dominant sign of the parameter updates across adapters is chosen for each parameter, and 3) Merge, in which the updates that match the elected sign are parameter-wise averaged. 
%
On the other hand, Task arithmetic leverages \textit{task vectors} that are the difference between a model's fine-tuned weights and its pre-trained weights, \textit{e.g.} LoRA adapters, through simple operations like addition, subtraction, and analogy-based reasoning. 
%
However, we empirically show in Sec~\ref{sec:exp} that neither merging method alone is perfect to address heterogeneous task merging requests.
\fi

\input{method}
\input{experiments}

\vspace{-0.1in}
\section{Conclusion}

We present FULM, a federated unlearning framework for LLMs, which enables continual and decentralized knowledge removal without centralized data aggregation. To address the challenge of asymmetric data access, where separate parties hold unlearning and retention data, FULM decouples unlearning and retention objectives and performs hierarchical task vector merging, which adapts to both near-iid and heterogeneous unlearning requests while preserving critical knowledge. 
Comprehensive experiments on WMDP, TOFU, and MUSE benchmarks show that FULM effectively removes undesirable content, maintains utility, and scales to heterogeneous real-world unlearning scenarios, offering a practical and privacy-preserving solution for LLM unlearning.

% \clearpage
\bibliographystyle{ACM-Reference-Format}
\bibliography{main}

\clearpage 
\appendix

\section{Algorithm Overview}

\begin{algorithm}[ht]
\caption{One-Round Hierarchical Federated Unlearning Merging}
\label{algo:merge}
\begin{algorithmic}[1]
\Require 
\Statex - Clients $\{1, 2, ..., K\}$ with private unlearning datasets $\{D_u^k\}$ and optional retention datasets $\{D_r^k\}$
\Statex - Server with global model parameters $\theta$ and access to pretraining data.

\vspace{0.5em}
\State \textbf{Broadcast:} Server transmits current foundation model $\theta$ to all clients
\For{each client $k \in [K]$ \textbf{in parallel}}
    \State Perform LoRA-based unlearning on $D_u^k$ to obtain adapter $\nabla \theta_u^k$
    \If{$D_r^k \neq \emptyset$}
        \State Optionally perform retention fine-tuning to obtain $\nabla \theta_r^k$
    \EndIf
    \State Return task adapters $\nabla \theta_u^k$ (and optionally $\nabla \theta_r^k$) to the server
\EndFor

\vspace{0.5em}
\State \textbf{Similarity-aware Clustering:} Cluster all received task adapters $\{\nabla \theta_{u(r)}^k\}$ into sets $\{\mathcal{C}_1, ..., \mathcal{C}_{|\mathcal{C}|}\}$ based on cosine similarity

\vspace{0.5em}
\For{each cluster $\mathcal{C}_k$}
    \State Intra-cluster merging via voting: $\nabla \Theta_k \leftarrow \mathcal{A}_\text{vote} \left( \{\nabla \theta_j \in \mathcal{C}_k\} \right)$
\EndFor

\vspace{0.5em}
\State \textbf{Inter-cluster Aggregation:} Merge cluster-level vectors:
\[
\nabla \Theta \leftarrow \mathcal{A}_\text{sum} \left( \{ \nabla \Theta_k \}_{k=1}^{|\mathcal{C}|} \right)
\]

\vspace{0.5em}
\State \textbf{Server Retention (Optional):} If applicable, server performs retention fine-tuning using pretraining subset to obtain $\nabla \theta_r^\text{server}$

\vspace{0.5em}
\State \textbf{Model Update:} Apply combined updates:
\[
\theta' = \theta + \nabla \Theta + \nabla \theta_r^\text{server}
\]

\State \Return Updated model parameters $\theta'$
\end{algorithmic}
\end{algorithm}

\section{Additional Experimental Details}
As shown in Table~\ref{tab:tofu1}, continued unlearning over epochs progressively degrades model performance on the retained set, which shares distributional characteristics with the forget set (i.e., fabricated character biographies in TOFU). In contrast, general knowledge domains such as Real Authors and Real World remain relatively stable, indicating that unlearning primarily affects semantically proximate content.

\begin{table}[htbp!]
\centering
\scriptsize
\caption{Impact of Continued Unlearning on Retention and Utility over Epochs (TOFU)}
\label{tab:tofu1}
\begin{tabular}{@{}lccccc@{}}
\toprule
Epoch  &  Real Authors$\uparrow$ &  Real World$\uparrow$ &  Retain$\uparrow$ &  Forget$\downarrow$ & Utility$\uparrow$\\
\midrule
1  & 71.20\% & 81.94\% & 89.93\% & 91.54\% & 58.89\%\\
2  & 69.70\% & 82.51\% & 89.30\% & 83.15\% & 58.77\%\\
3  & 63.85\% & 81.65\% & 83.33\% & 71.03\% & 57.57\%\\
4  & 59.35\% & 82.68\% & 69.92\% & 58.53\% & 55.49\%\\
5  & 58.35\% & 79.94\% & 66.08\% & 55.10\% & 54.44\%\\
\bottomrule
\end{tabular}
\end{table}

As shown in Table~\ref{tab:tofu2}, increasing the size of the forgetting set leads to broader degradation in both model utility and retention performance. This highlights a trade-off between unlearning strength and knowledge preservation: larger forget sets impose greater disruption on the model's internal representations, thereby weakening its ability to retain unrelated or general knowledge.

\begin{table}[htbp!]
\centering
\scriptsize
\caption{Impact of Forgetting Set Size on TOFU Dataset}
\label{tab:tofu2}
\begin{tabular}{@{}lccccc@{}}
\toprule
Forget Set Size & Model Utility↑ &  Real Authors↑ &  Real World↑ &  Retain↑ & Forget↓ \\
\midrule
25\% of dataset & 55.97\% & 72.25\% & 84.62\% & 72.34\% & 71.23\% \\
5\% of dataset  & 57.16\% & 69.42\% & 80.80\% & 82.06\% & 69.95\% \\
\bottomrule
\end{tabular}
\end{table}

% \subsection{Adapter Training Objective for WMDP Unlearning}

% To unlearn hazardous knowledge from WMDP in the domains of biosafety and cybersecurity (1,273 biology questions and 1,987 cybersecurity questions), we fine-tune an adapter using the RMU (Representation Misdirection for Unlearning) method. The training objective consists of a single \textit{forget loss}, which alters the model's internal representations on hazardous inputs.

% \paragraph{Forget Loss.} Let $M_{\text{updated}}(\cdot)$ denote the hidden activations of the adapter-augmented model at a selected layer $\ell$, and let $\mathbf{u} \in \mathbb{R}^d$ be a random unit vector sampled uniformly from $[0, 1]^d$. For a forget dataset $D_{\text{forget}}$, the forget loss encourages the model to deviate from its original representations in a random direction:

% \begin{equation}
% \mathcal{L}_{\text{forget}} = \mathbb{E}_{x_f \sim D_{\text{forget}}} \left[ \frac{1}{L_f} \sum_{\text{token } t \in x_f} \left\| M_{\text{updated}}(t) - c \cdot \mathbf{u} \right\|_2^2 \right]
% \end{equation}

% where $L_f$ is the number of tokens in the hazardous input $x_f$, and $c$ is a hyperparameter controlling the scaling of the perturbation.

% \paragraph{Training Procedure.} During adapter training, we interleave gradient updates from biosecurity and cybersecurity examples in $D_{\text{forget}}$, and minimize $\mathcal{L}_{\text{forget}}$. We update only the adapter layers (at the down-projection linear layers of the fifth to seventh layers) to perform unlearning while keeping the backbone model frozen.

\subsection{Adapter Training Objectives}\label{Adapter}
We apply different adapter finetuning objectives for unlearning across three benchmarks: \textbf{WMDP}, \textbf{TOFU}, and \textbf{MUSE}, each targeting distinct types of sensitive knowledge.

\textbf{WMDP (Biosecurity and Cybersecurity).}
To unlearn hazardous knowledge in biosafety and cybersecurity domains (1,273 biology and 1,987 cybersecurity questions), we employ the RMU (Representation Misdirection for Unlearning) method. The training objective consists of a forget loss that perturbs model activations on hazardous inputs $x_f \sim D_{\text{forget}}$ at an intermediate layer $\ell$:

\begin{equation}
\mathcal{L}_{\text{forget}} = \mathbb{E}_{x_f \sim D_{\text{forget}}} \left[ \frac{1}{L_f} \sum_{\text{token } t \in x_f} \left\| M_{\text{updated}}(t) - c \cdot \mathbf{u} \right\|_2^2 \right]
\end{equation}

where $\mathbf{u} \sim \text{Uniform}([0, 1]^d)$ is a random unit vector, $c$ is a hyperparameter, and $M_{\text{updated}}(\cdot)$ denotes the model activations at layer $\ell$.

\textit{Training Procedure.} During adapter training, we interleave gradient updates from biosecurity and cybersecurity examples in $D_{\text{forget}}$, and minimize $\mathcal{L}_{\text{forget}}$. We update only the adapter parameters in the down-projection linear layers of Transformer layers 5 through 7, while keeping the backbone model frozen to ensure efficient and localized unlearning.

\textbf{TOFU and MUSE (Gradient Ascent Unlearning).}
For the TOFU and MUSE benchmarks, we apply a gradient ascent approach to encourage the model to forget specific content by maximizing the standard training loss on the forget dataset $x_f \sim D_{\text{forget}}$:

\begin{equation}
\mathcal{L}_{\text{forget}} = \mathbb{E}_{x_f \sim D_{\text{forget}}} \left[ \ell(x_f, w) \right]
\end{equation}

where $\ell(x_f, w)$ is the loss function of the model (e.g., cross-entropy) on input $x_f$ with weights $w$.

Although both benchmarks share this objective, they differ in data format:
\begin{itemize}
  \item \textbf{TOFU} contains fabricated personal information presented in the form of question-answer (Q\&A) pairs, such as ``What is Alice Zhang’s phone number?''.
  \item \textbf{MUSE} consists of long-form text passages extracted from copyrighted books, such as \emph{Harry Potter and the Goblet of Fire}.
\end{itemize}

In both cases, we train the adapter to disrupt the model’s ability to reproduce the target content.

\subsection{Full experimental results of Intra- and Inter-Cluster Merging.}
\label{full}
Table~\ref{tab:wmdp_iid_full} and Table~\ref{tab:wmdp_noniid_full} present the full results for both stages of our hierarchical unlearning merging process.

\textbf{Step 1: Intra-Cluster Merging.} We include individual unlearning vectors ($Cyber_i$, $Bio_i$) alongside the aggregated results (AVG, TIES, SUM).

\begin{table}[htbp]
\centering
\scriptsize
\vspace{-0.1in}\caption{Intra-Cluster Merging (Step 1)} \vspace{-0.1in}
\label{tab:wmdp_iid_full}
\begin{minipage}[t]{0.47\linewidth}
\centering
\caption*{\textbf{(a) WMDP-Cyber}}\vspace{-0.1in}
\begin{adjustbox}{max width=\linewidth}
\begin{tabular}{@{}lccc@{}}
\toprule
Method & Bio$\downarrow$ & Cyber$\downarrow$ & Util.$\uparrow$ \\
\midrule
Pretrained & 63.71\% & 44.00\% & 58.19\% \\
\arrayrulecolor{gray!50}\midrule
\arrayrulecolor{black}
$Cyber_1$ & 65.20\% & 27.48\% & 57.95\% \\
$Cyber_2$ & 65.28\% & 26.62\% & 58.34\% \\
$Cyber_3$ & 65.12\% & 26.37\% & 52.07\% \\
\arrayrulecolor{gray!50}\midrule
\arrayrulecolor{black}
AVG & 65.44\% & 28.52\% & 58.74\% \\
\textbf{TIES} & 65.44\% & \textbf{26.02\%} & \textbf{58.91}\% \\
SUM & \textbf{24.98}\% & 26.62\% & 23.90\% \\
\bottomrule
\end{tabular}
\end{adjustbox}
\end{minipage}
\hspace{0.03\linewidth}
\begin{minipage}[t]{0.47\linewidth}
\centering
\caption*{\textbf{(b) WMDP-Bio}}\vspace{-0.1in}
\begin{adjustbox}{max width=\linewidth}
\begin{tabular}{@{}lccc@{}}
\toprule
Method & Bio$\downarrow$ & Cyber$\downarrow$ & Util.$\uparrow$ \\
\midrule
Pretrained & 63.71\% & 44.00\% & 58.19\% \\
\arrayrulecolor{gray!50}\midrule
\arrayrulecolor{black}
$Bio_1$ & 33.15\% & 44.24\% & 57.64\% \\
$Bio_2$ & 32.13\% & 44.24\% & 58.78\% \\
$Bio_3$ & 63.94\% & 44.74\% & 58.47\% \\
\arrayrulecolor{gray!50}\midrule
\arrayrulecolor{black}
AVG & 47.36\% & 44.22\% & 58.02\% \\
\textbf{TIES} & {41.32\%} & 44.29\% & 57.95\% \\
SUM & \textbf{24.98}\% & \textbf{32.51}\% & 52.07\% \\
\bottomrule
\end{tabular}
\end{adjustbox}
\end{minipage}\vspace{-0.1in}
\end{table}

\textbf{Step 2: Inter-Cluster Merging.} After intra-cluster merging, we obtain three centroid vectors (Cyber, Bio, HP) and apply merging across non-iid domains.

\begin{table}[htbp]
\centering
\scriptsize
\caption{Inter-Cluster Merging (Step 2)}\vspace{-0.1in}
\label{tab:wmdp_noniid_full}
\begin{tabular}{@{}lcccc@{}}
\toprule
Unlearning Tasks & Biology$\downarrow$ & Cyber Security$\downarrow$ & Harry Potter$\downarrow$ & Model Utility$\uparrow$ \\
\midrule
Pretrained Model & 63.71\% & 44.0\% & 49.87\% & 58.19\% \\
\arrayrulecolor{gray!50}\midrule
\arrayrulecolor{black}
Cyber  & 65.44\% & 26.02\% & 47.17\%& 58.91\% \\
Bio  & 41.32\% & 44.29\% & 47.32\%& 57.95\% \\
Harry Potter (HP)  & 63.36\% & 43.81\% & 33.54\% & 57.89\%\\
\arrayrulecolor{gray!50}\midrule
\arrayrulecolor{black}
Avg  & 63.63\% & 43.03\% & 46.36\% & \textbf{58.39}\% \\
TIES  & 62.45\% & 41.87\% & 46.71\% & 58.28\% \\
\textbf{SUM }& \textbf{39.05\%} & \textbf{25.82\%} & \textbf{34.48\%}& 57.25\% \\

\bottomrule
\end{tabular}
\end{table}

\end{document}

%% file: method.tex
% \vspace{-0.1in}
\section{Method}
\noindent \textbf{Problem Setting:} We consider a federated unlearning framework where a central server hosts a foundation model (\textit{e.g.}, an open-sourced LLM on the Hugging Face platform) and receives multiple unlearning requests from decentralized clients. 
Each client $k \in [K]$ has \textit{white-box} access to LLM parameters $\theta$ and a private unlearning dataset $D^k_u$ containing knowledge to be forgotten. Optionally, the client may obtain a retention dataset $D^k_r$ of information to preserve, although it is often absent with $D^k_r = \emptyset$. 
The server maintains access to pretraining data, which partially overlaps with client data but cannot be shared due to privacy constraints.

\subsection{Split-and-Merge: Decoupled Unlearning and Retenion via Dual Task adapters}

%To address the challenge of asymmetric data access and balance the goals of forgetting and retention, we adopt a split-and-merge strategy. Specifically, the unlearning and retaining datasets are handled independently: while each client performs unlearning locally, the server can reinforce the model on a private retention dataset to produce a distinct task vector that captures preservation-oriented updates. Since this vector is directionally dissimilar to most unlearning updates, similar to a non-iid task, it is merged via arithmetic addition alongside other non-iid vectors. 
%This modular integration allows the model to support diverse deletion requests while maintaining overall utility.

To address asymmetric data access while maintaining balanced forgetting and preservation, we first propose to decouple unlearning and retention into independent adaptation tasks.
Our framework encompasses two phases: a \textit{split} phase where unlearning and retention are handled independently, and \textit{merge} phase where the resulting adapters are aggregated at the server. 

Specifically, clients first perform unlearning locally on their private datasets $D^k_u$ using standard unlearning objectives, such as gradient ascent~\cite{pmlr-v132-neel21a}.
Whereas, retention adapters can be generated either by clients with available retention data $D^k_r$, or by the server using a lightweight subset of pretraining data.
Without losing generality, we assume that each adaptation is encapsulated via LoRA fine-tuning, although our methods can be extended to varying adapter architectures that capture task-specific parameter updates.
Empirical results (Section \ref{sec:exp}) show that this decoupled approach outperforms traditional unlearning with a dual objective as in Eq~\ref{eq:unlearn-dual}, especially when the unlearning and retention datasets are resemblant and thus conflicting, such as in structured \textit{entry-wise element} unlearning scenarios~\cite{maini2024tofutaskfictitiousunlearning}.

%\vspace{-0.1in}
\subsection{Multi-Task Unlearning Transfer via Hierarchical Federated Merging}
Once the LLM server collects multiple task adapters for unlearning and retention, the core challenge resides in an effective merging approach to balance two complementary tradeoffs: the \textit{inter-domain} unlearning interference, and the \textit{intra-domain} interference between potentially conflicting unlearning and retaining task objectives.

Let $\nabla \theta_{u(r)}^i$ denote the unlearning (or retaining) task adapter trained on dataset $\mathcal{D}_i$. The above interferences mainly stem from two aspects: (1) the underlying data distribution manifested by $\mathcal{D}_i$, and (2) the task objective $\mathcal{L}(D_i; \theta)$ of either forgetting or preserving knowledge from such dataset.
We take a hierarchical, data-free merging approach to tackle task vectors' heterogeneity induced by both their data and learning objectives, without presuming the LLM server to access proxy data for merging.

\noindent \textbf{Similarity-aware Adapter Clustering}:
The LLM server first clusters task vectors according to their parameter distribution similarity. Our rationale is that task adapters trained with similar objectives $\mathcal{L}(\cdot)$ and data distributions $\mathcal{D}_i$ produce parameter updates induced by $\nabla \mathcal{L}(D_i; \theta)$, that adapt the model toward similar directions, resulting in positively correlated perturbations, and vice versa.

\begin{figure}[htbp!]
\centering
\includegraphics[width=0.4\textwidth]{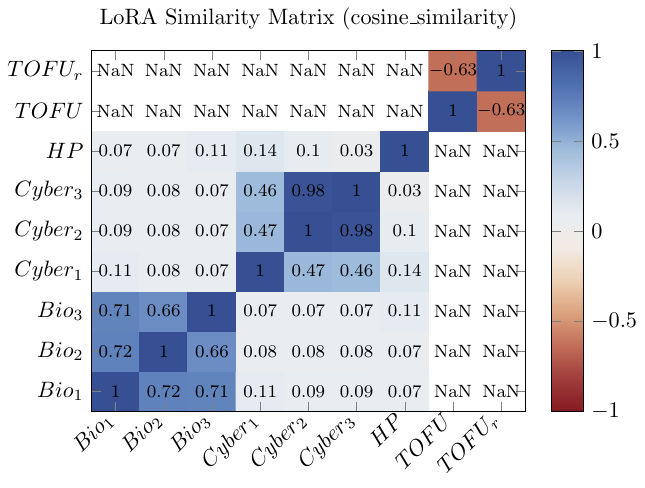}
\caption{Cosine similarity across task vectors, where vectors from near-iid sources exhibit high similarity, and those from heterogeneous domains are nearly orthogonal. Task vectors trained for retention ($TOFU_r$) show negative correlation to their unlearning counterparts ($TOFU$).}
\label{fig:lora_similarity_extended}
\vspace{-0.1in}
\end{figure}

Without losing generality, we use \textsc{cosine} similarity as the task adapter correlation metric. The results of our empirical study in Figure \ref{fig:lora_similarity_extended} confirm our design, in that both the data distribution and the learning objectives influence the similarity patterns of the task vector. Specifically, (i) unlearning task adapters trained on orthogonal datasets $D_i$ and $D_j$ produce orthogonal parameter shifts with near-zero cosine similarity. Meanwhile, (ii) unlearning adapters trained on near-iid datasets exhibit a strong positive correlation in their parameter updates. Interestingly, (iii) unlearning and retention adapters $\theta_u^i$ and $\theta_r^j$ trained on disjoint but near-iid data $D_i$ and $D_j$ demonstrate negatively correlated similarity that reflects their contrastive objectives.

\noindent \textbf{Hierarchical Merging Strategies:} Following task vector clustering via cosine similarity, we propose a two-step hierarchical merging strategy:

\paragraph{Step 1: Intra-Cluster Merging.}
We first define a similarity threshold $\xi > 0$ to group adapters with high cosine similarity, which correspond to near-iid data and aligned objectives.
Within each cluster $\mathcal{C}_k$, we apply a voting-based merging strategy such as TIES~\cite{yadav2023tiesmergingresolvinginterferencemerging}, which selects dominant parameter directions while minimizing destructive interference. Compared to task arithmetic methods (\textit{e.g.} addition), voting-based merging avoids over-amplifying similar updates and lead to more stable representations for tightly aligned tasks (\Eason{see Section~\ref{sec:empirical}}). \J{For parameter-efficient adapters like LoRA, similarity is computed after recovering low-rank representations in the original parameter space. Thus, our method is agnostic to specific adapter architectures, so long as they are compatible with numerical operations on the same LLM backbone. For instance,  adapters can be learned with LoRA that target different linear layer combinations (\textit{e.g.} selective \textit{vs}. all attention layers).}

\paragraph{Step 2: Inter-Cluster Merging.}
We then treat the merged cluster-level adapters as a set $\mathcal{C} = \{\nabla \Theta_k\}_{k=1}^{|\mathcal{C}|}$, where each $\nabla \Theta_k$ represents a merged vector from one cluster. Since these vectors capture distinct information domains, we use task arithmetic addition to preserve all unique update information without dilution.
The complete hierarchical merging process is formalized as: 
\begin{align*} 
\nabla \Theta \leftarrow \mathcal{A}_\text{sum}\Big( \big\{\nabla \Theta_k \in \mathcal{C}  | \Theta_k \leftarrow  \mathcal{A}_\text{vote} ( \{\nabla \theta_j \in \mathcal{C}_k \}_{j=1}^{|\mathcal{C}_k|}) \big\}_{k=1}^{|\mathcal{C}|} \Big),
\end{align*}

where $\mathcal{A}_{\text{vote}}$ represents voting-based intra-cluster aggregation and $\mathcal{A}_{\text{sum}}$ denotes summation across clusters. This approach eliminates the need to manage multiple task-specific adapters and supports scalability for continual and dynamic unlearning requests. 
\J{The overall process is summarized in Algorithm~\ref{algo:merge} of Appendix.}

% Here, $\Delta\theta_{iid}^{i}$ represents task vectors from iid data, $\Delta \theta_j$ denotes vectors from non-iid data, and $\Delta\theta_{retain}$ encapsulates the retaining task vector to balance retaining and forgetting tasks. This unified framework ensures effective integration of diverse unlearning needs while simultaneously preserving model utility and privacy across federated learning environments. 

%% file: experiments.tex
\section{Experiments}\label{sec:exp}
\subsection{Experimental Setup}

\noindent \textbf{Datasets}: We conducted experiments on three representative unlearning benchmarks: (1) \textbf{WMDP}~\cite{yao2024large}, for which we employed two domain-specific forgetting sets, Biosecurity and Cybersecurity, that contain sensitive content sourced from PubMed and GitHub;
(2)  \textbf{TOFU}~\cite{maini2024tofutaskfictitiousunlearning}, which is a QA benchmark featuring fictional authors as the unlearning target and real-world factual QA pairs as the retention set; and 
(3) \textbf{MUSE}~\cite{nguyen2024surveymachineunlearning}, from which we selected a subset of  the \textit{Harry Potter (HP)} series,  using contents from \textit{Harry Potter and the Goblet of Fire} as the forgetting set.

\noindent \textbf{Evaluation Metrics.}  
We primarily measured the \textit{Forget}$\downarrow$ performance using the forgetting set to assess how well target content is removed, and the model \textit{Utility}$\uparrow$ as the retention performance, which is evaluated on MMLU~\cite{hendrycks2021measuringmassivemultitasklanguage} benchmark that contains 14,079 questions across 57 tasks.
Additionally, for TOUFU-related tasks, we added three more metrics: \textit{Retain} $\uparrow$, \textit{Real Authors$\uparrow$}, and \textit{Real World}$\uparrow$,  all measured using ROUGE, a standard metric that computes the textual overlap between model outputs and reference texts. 
Specifically, \textit{Retain}$\uparrow$ on a held-out retaining set measures the preservation of useful knowledge; and \textit{Real Authors}$\uparrow$ / \textit{Real World}$\uparrow$ refer to factual QA subsets in the TOFU dataset not intended for removal. 
%
%These ROUGE-based metrics together capture both unlearning success and unintended collateral forgetting. 
%
\J{We eventually report an averaged, balanced performance metric:}
$\text{Overall}=\frac{1}{|\mathbf{M}|}\Big[{\sum_i{m^i_\text{r}}} +\sum_j(1 - m^j_\text{u})   \Big]$, 
%where we reverse a metric $m^j_u$ indicating unlearning performance and reserve the metric $m^i_r$ indicating its preserving performance (\textit{e.g.} utility) on desirable knowledge.
%
where $m^i_r$ measures retention (\textit{e.g.} utility on desirable knowledge), and $m^j_u$  measures unlearning, reversed to reflect forgetting effectiveness.

\noindent \textbf{Baselines.} 
For evaluating \textit{merging} performance, we compared our method against the following baselines: (1) the \textit{Avg} that computes the parameter-wise average of all  task vectors, (2) \textit{SUM} which performs direct arithmetic addition, (3) \textit{TIES}~\cite{yadav2023tiesmergingresolvinginterferencemerging}, a voting-based gradient merging strategy, and (4) KNOT~\cite{stoica2024modelmergingsvdtie} that aligns LoRA adapters on a low-dimensional parameter subspace via joint SVD-based transformations to improve merge quality. 
To compare decoupled versus joint unlearning, we adopt \textit{Gradient Difference (GD)} from \cite{maini2024tofutaskfictitiousunlearning} as a baseline, which is a joint optimization approach by minimizing the composite loss of Equation \ref{eq:unlearn-dual}.

\noindent \textbf{Decentralized Unlearning Scenarios:} We adopted the Zephyr-7B  model as our base LLM. All evaluated task adapters were learned with LoRA with a rank $r = 16$ and a scaling factor of $32$. For the TOFU and MUSE datasets, we applied LoRA to all linear layers, while for WMDP, we applied LoRA only to the down-projection linear layers of the fifth to seventh layers. We evaluated two FL settings: (1) a \textbf{\textit{near-iid}} setup using TOFU datasets, where the unlearning dataset was evenly split across five clients, along with a retention adapter trained on a separate dataset that was unavailable to clients; and (2) a \textbf{\textit{heterogeneous}} setup with seven unlearning adapters trained on diverse domains, with three on \textit{WMDP-Bio}, three on \textit{WMDP-Cyber}, where the WMDP-Cyber and WMDP-Bio forgetting sets were randomly partitioned into three subsets each, and one on \textit{HP}.
For both settings, all unlearning adapters were learned with only unlearning data.  
\Eason{See Appendix \ref{Adapter} for details.}

% Two specially curated datasets are utilized:

% \textbf{TOFU Dataset:} Contains QA pairs related to fictitious authors (Forget Set) and other knowledge types (Retain Set). It evaluates unlearning efficacy under realistic conditions.

% \textbf{WMDP Dataset:} WMDP benchmark includes a forgetting corpus comprising 1,000 samples (21.4MB) for (WMDP-Cyber) and 24,453 samples (712.96MB) for WMDP-Bio, which are collected from PubMed and Github.

% It evaluates unlearning strategies targeting hazardous knowledge in Biosecurity and Cybersecurity. Categorized into general, expert-level, and hazardous knowledge, this dataset assesses selective removal of harmful information.

% Following WMDP, we employed WikiText-2-raw-v1\cite{wikitext} dataset (36,718 samples) as retaining dataset, which consists of verified Good and Featured Wikipedia articles.

\vspace{-0.1in}
\subsection{Performance Evaluation} \label{sec:empirical}

\subsubsection{Effects of Hierarchical Merging:}
Table \ref{table:main-iid} and Table \ref{table:main-hetero} summarize the performance of FULM and baseline merging methods under near-iid and heterogeneous setups, respectively. Across both scenarios, our hierarchical merging strategy consistently achieves the most balanced trade-off between unlearning and retention.

In the \textbf{\textit{near-iid setting}} (Table \ref{table:main-iid}), where merging was performed on 5 client-provided unlearning adapter $\nabla \theta_u^i$ and 1-server provided retention adapter $\theta_r$, non-hierarchical baselines struggled to maintain balance. For example, the  SUM merge amplifies unlearning but severely degrades utility (56.92\%$\uparrow$), compared to FULM (70.62\%$\uparrow$). Meanwhile, TIES and AVG disrupt retention effects and lead to impaired performance on desirable knowledge categories like \textit{Real Authors}/\textit{Real World}. \J{The KNOT baseline that leverages SVD prior to adapter merging shows negligible difference from other baselines.} 

For the \textbf{\textit{heterogeneous}} setting (Table \ref{table:main-hetero}), non-hierarchical merging lead to severe cross-domain interference and ineffective unlearning. While SUM merging achieves comparable unlearning to FULM, it dramatically sacrifices model utility. FULM maintains strong forgetting while effectively preserving critical knowledge. \J{\textit{KNOT} merging is excluded from evaluation of heterogeneous setting, as it requires unified LoRA architectures for SVD decomposition.}

%\vspace{-0.1in}
\begin{table}[htbp!]
\centering
\scriptsize
\vspace{-0.1in} 
\caption{\Eason{Merging Near-iid Unlearning Domains.} \label{table:main-iid}} 
\vspace{-0.1in}
\begin{tabular}{@{}lcccccc@{}}
\toprule
Method & Real Authors$\uparrow$ & Real World$\uparrow$ & Retain$\uparrow$ & Forget$\downarrow$ & Utility$\uparrow$ & \textbf{Overall}$\uparrow$ \\
\midrule
Ties & 50.38\% & 59.05\% & 48.94\% & 50.67\% & 57.42\% & 53.02\% \\
SUM & 87.58\% & \textbf{78.95}\% & 40.38\% & \textbf{39.17}\% & 56.92\% & 64.93\% \\
AVG & 62.71\% & 61.79\% & 49.66\% & 49.90\% & 63.29\% & 57.51\% \\
KnOT & 52.37\% & 61.84\% & 45.59\% & 48.12\% & 56.25\% & 66.21\%\\
FULM (proposed) & \textbf{99.06}\% & 78.35\% & \textbf{64.48}\% & 51.84\% & \textbf{70.62}\% & \textbf{72.13}\% \\
\bottomrule
\end{tabular}
\end{table}

\vspace{-0.1in}
\begin{table}[htbp]
\centering
\scriptsize
\vspace{-0.1in}\caption{\Eason{Merging Heterogeneous Unlearning Domains.} \label{table:main-hetero}} \vspace{-0.1in}
\label{tab:main2}
\begin{tabular}{@{}lccccc@{}}
\toprule
Unlearning Tasks & Biology$\downarrow$ & Cyber Security$\downarrow$ & Harry Potter$\downarrow$ & Model Utility$\uparrow$ & \textbf{Overall}$\uparrow$ \\
\midrule
Avg All Vectors & 59.85\% & 38.42\% & 41.17\% & 56.79\% & 54.34\% \\
TIES All Vectors & 62.45\% & 41.87\% & 46.71\% & \textbf{58.28}\% & 51.81\% \\
SUM All Vectors & \textbf{27.46}\% & 34.47\% & \textbf{29.31}\% & 30.42\% & 59.80\% \\
FULM (proposed) & 39.05\% & \textbf{25.82}\% & 34.48\% & 57.25\% & \textbf{64.48}\% \\

\bottomrule
\end{tabular}
\end{table}

\subsubsection{Effects of Decoupled Unlearning and Retention :}
%Experiments on the TOFU dataset provide deeper insights into the dynamics of forgetting and retention. Table~\ref{tab:tofu3} compares our split-and-merge strategy with other baselines. Among all methods, the proposed method achieves the best balance between forgetting and retaining, attaining effective unlearning performance while maximizing the preservation of the model’s general capability. These results indicate that decoupling the unlearning and retention objectives, followed by hierarchical task vector merging, is more effective than applying a uniform merge strategy to all task vectors or jointly optimizing them within a single training loop. This also makes our method well-suited for scenarios with asymmetric data access, where unlearning and retention data are held by different parties and cannot be directly combined.
As shown in Table~\ref{tab:tofu3}, our split-and-merge strategy achieves optimal balance when merging one unlearning and one retention adapter. Notably, our decoupled approach outperforms \textit{GD} (joint training on combined datasets) and demonstrate that separate optimization followed by task-arithmetic merging is more effective than joint optimization especially when data conflicts. This also makes our method suitable for asymmetric data access  where different parties hold unlearning and retention data.

\begin{table}[htbp!]
\centering
\scriptsize
\caption{Effects of merging two unlearning and retention adapters on the TOFU dataset. GD represents joint training.}\vspace{-0.1in}
\label{tab:tofu3}
\begin{tabular}{@{}lcccccc@{}}
\toprule
Method & Real Authors$\uparrow$ & Real World$\uparrow$ & Retain$\uparrow$ & Forget$\downarrow$ & Utility$\uparrow$ & \textbf{Overall}$\uparrow$ \\
\midrule
GD (centralized) & 55.27\% & 74.93\% & \textbf{89.62}\% & 53.57\% & 67.16\% & 66.68\% \\
KnOT & \textbf{86.83}\% & 75.78\% & 48.94\% & 50.19\% & 69.97\% & 66.27\% \\
FULM (SUM) & 86.13\% & \textbf{75.85}\% & 48.37\% & \textbf{20.79}\% & \textbf{70.62}\% & \textbf{72.04}\% \\
\bottomrule
\end{tabular}
\end{table}

\subsubsection{Breaking Down Analysis of Hierarchical Unlearning Merging}

\noindent \textbf{Step1}: Intra-Cluster Merging on Near-iid Task Vectors:
We investigated different unlearning merging strategies for adapters within a cluster.
%, , including AVG, TIES, and SUM.
%
%outperforms other merging methods and even other individual task vectors. It 
As shown in Table~\ref{tab:wmdp_iid}, TIES achieves stronger unlearning (e.g., Cyber $26.02\%$ and Bio $41.32\%$) while preserving utility ($58.91\%$ and $57.95\%$), especially outperforming SUM, which severely degrades retention utility, as it accumulates update magnitudes across unlearning tasks, leading to overly large activations that compromise model general usability.
These  indicate that voting-based merging (TIES) is suitable for near-iid task vectors by mitigating mutual interference.while promoting generalization. See Table \ref{tab:wmdp_iid_full} in Appendix \ref{full} for full results.
% across similar unlearning tasks.%, achieving better forgetting without sacrificing utility.

\begin{table}[htbp]
\centering
\scriptsize
\vspace{-0.1in}\caption{Intra-Cluster Merging (Step 1)} \vspace{-0.1in}
\label{tab:wmdp_iid}
\begin{minipage}[t]{0.47\linewidth}
\centering
\caption*{\textbf{(a) WMDP-Cyber}}\vspace{-0.1in}
\label{tab:wmdp_iid_a}
\begin{adjustbox}{max width=\linewidth}
\begin{tabular}{@{}lcccc@{}}
\toprule
Method & Bio$\uparrow$ & Cyber$\downarrow$ & Util.$\uparrow$ & Overall$\uparrow$ \\
\midrule
Pretrained & 63.71\% & 44.00\% & 58.19\% & 59.30\% \\
\arrayrulecolor{gray!50}\midrule
\arrayrulecolor{black}
AVG & 65.44\% & 28.52\% & 58.74\% & 65.22\% \\
\textbf{TIES} & \textbf{65.44\%} & \textbf{26.02\%} & \textbf{58.91\%} & \textbf{66.11\%} \\
SUM & 24.98\% & 26.62\% & 23.90\% & 40.75\% \\
\bottomrule
\end{tabular}
\end{adjustbox}
\end{minipage}
\hspace{0.03\linewidth}
\begin{minipage}[t]{0.47\linewidth}
\centering
\caption*{\textbf{(b) WMDP-Bio}}\vspace{-0.1in}
\label{tab:wmdp_iid_b}
\begin{adjustbox}{max width=\linewidth}
\begin{tabular}{@{}lcccc@{}}
\toprule
Method & Bio$\downarrow$ & Cyber$\uparrow$ & Util.$\uparrow$ & Overall$\uparrow$ \\
\midrule
Pretrained & 63.71\% & 44.00\% & \textbf{58.19\%} & 46.16\% \\
\arrayrulecolor{gray!50}\midrule
\arrayrulecolor{black}
AVG & 47.36\% & 44.22\% & 58.02\% & 51.63\% \\
\textbf{TIES} & 41.32\% & \textbf{44.29\%} & 57.95\% & \textbf{53.64\%} \\
SUM & \textbf{24.98\%} & 32.51\% & 52.07\% & 53.20\% \\
\bottomrule
\end{tabular}
\end{adjustbox}
\end{minipage}\vspace{-0.1in}
\end{table}

\noindent \textbf{Step2}: Merging non-IID Task Vectors:
%After applying TIES to merge task vectors within each cluster, we proceed with the second step of hierarchical aggregation for non-IID aggregation. We obtained aggregated task vectors for WMDP-Cyber and WMDP-Bio, as well as forgetting task vectors trained on MUSE-HP, representing heterogeneous domains. Table~\ref{tab:wmdp_noniid} shows that SUM merging method achieves the best forgetting across all domains while maintaining competitive utility ($57.25\%$). This indicates that for non-IID task vectors, direct summation effectively preserves independent forgetting effects without interference, making it a simple yet effective aggregation strategy in heterogeneous scenarios.
After applying TIES within each cluster, we obtain three centroid adaptors: WMDP-Cyber, WMDP-Bio, and MUSE-HP. We then perform the second-stage merging across these non-iid domains. As shown in Table~\ref{tab:wmdp_noniid}, the SUM method delivers the strongest forgetting across all domains while maintaining competitive utility (57.25\%). This indicates that, in heterogeneous scenarios, direct summation preserves domain-specific forgetting with minimal interference as a simple yet effective aggregation strategy. Full results are shown in Table \ref{tab:wmdp_noniid_full} in Appendix \ref{full}.

\vspace{-0.1in}
\begin{table}[htbp]
\centering
\scriptsize
\caption{Inter-Cluster Merging (Step 2)}\vspace{-0.1in}
\label{tab:wmdp_noniid}
\begin{tabular}{@{}lccccc@{}}
\toprule
Unlearning Tasks & Biology$\downarrow$ & Cyber Security$\downarrow$ & Harry Potter$\downarrow$ & Model Utility$\uparrow$ & Overall$\uparrow$ \\
\midrule
Pretrained Model & 63.71\% & 44.00\% & 49.87\% & 58.19\% & 50.15\% \\
\arrayrulecolor{gray!50}\midrule
\arrayrulecolor{black}
Avg & 63.63\% & 43.03\% & 46.36\% & \textbf{58.39\%} & 51.34\% \\
TIES & 62.45\% & 41.87\% & 46.71\% & 58.28\% & 51.81\% \\
\textbf{SUM} & \textbf{39.05\%} & \textbf{25.82\%} & \textbf{34.48\%} & 57.25\% & \textbf{64.48\%} \\

\bottomrule
\end{tabular}
\end{table}